\documentclass{article}
\usepackage{arxiv}
\usepackage[utf8]{inputenc} 
\usepackage[T1]{fontenc}    
\usepackage{hyperref}       
\usepackage{url}  
\usepackage{algorithmic}
\usepackage{algorithm}
\usepackage{booktabs}       
\usepackage{amsfonts}  
\usepackage{amsmath}
\usepackage{nicefrac}       
\usepackage{microtype}     
\usepackage{lipsum}		
\usepackage{graphicx}
\usepackage[square, numbers, comma, sort&compress]{natbib} 
\usepackage{doi}
\title{WideCaps: A Wide Attention based Capsule Network for Image Classification}
\author{S J Pawan $^{a}$ \quad Rishi Sharma$^{a}$\quad  Hemanth Sai Ram Reddy$^{a}$\quad M Vani$^{a}$\quad Jeny Rajan$^{a}$
\\
\\
$^{a}$Department of Computer Science and Engineering\\
National Institute of Technology Karnataka\\ Surathkal, India

}
\hypersetup{
pdftitle={A Wide Attention based Capsule Network for Image Classification},
pdfsubject={q-bio.NC, q-bio.QM},
pdfauthor={S J Pawan, Rishi Sharma, Hemanth Sai Ram Reddy, M Vani, Jeny Rajan },
pdfkeywords={Capsule Network, Convolutional Neural Network, Image Classification},
}

\begin{document}
\maketitle

\begin{abstract}
The capsule network is a distinct and promising segment of the neural network family that drew attention due to its unique ability to maintain the equivariance property by preserving the spatial relationship amongst the features. The capsule network has attained unprecedented success over image classification tasks with datasets such as MNIST and affNIST by encoding the characteristic features into the capsules and building the parse-tree structure. However, on the datasets involving complex foreground and background regions such as CIFAR-10, the performance of the capsule network is sub-optimal due to its naive data routing policy and incompetence towards extracting complex features. This paper proposes a new design strategy for capsule network architecture for efficiently dealing with complex images. The proposed method incorporates wide bottleneck residual modules and the Squeeze and Excitation attention blocks upheld by the modified FM routing algorithm to address the defined problem. A wide bottleneck residual module facilitates extracting complex features followed by the squeeze and excitation attention block to enable channel-wise attention by suppressing the trivial features. This setup allows channel inter-dependencies at almost no computational cost, thereby enhancing the representation ability of capsules on complex images. We extensively evaluate the performance of the proposed model on three publicly available datasets, namely  CIFAR-10, Fashion MNIST, and SVHN, to outperform the top-5 performance on CIFAR-10 and Fashion MNIST with highly competitive performance on the SVHN dataset.
\end{abstract}


\section{Introduction}
Ever since the rise of Convolutional Neural Networks or CNNs, it has become a gold standard for addressing almost every challenging problem related to computer vision. Of late, CNNs have become more effective and attained near-human-level performance on discriminative and generative tasks.  We can observe the success march of CNNs from LeNet-5 \cite{lecun1998gradient}, which was the first CNN-based neural network model for recognizing the handwritten and machine-printed characters, followed by the AlexNet \cite{krizhevsky2012imagenet}, which won the annual ImageNet Large Scale Visual Recognition Challenge-2012 (ILSVRC) \cite{ILSVRC15} by a large margin. After this, a plethora of CNN architectures emerged in the literature, with Overfeat \cite{sermanet2013overfeat}, VGG \cite{simonyan2014very}, GoogleNet \cite{szegedy2015going}, and Exception \cite{chollet2017xception} being the prominent ones. These models majorly focussed on increasing the depth and width of the architecture with extensive trainable parameters to improve the performance. However, over the period, researchers have discovered that the naive stacking of layers would result in the complications such as vanishing gradient and accuracy saturation resulting in poor performance. To subdue, the concept of residual connection or ResNet was introduced \cite{he2016deep}, which used identity skip connections to facilitate an easy gradient flow throughout the network. Subsequently, the literature has witnessed the surge of variants of residual network-based architectures such as DenseNet \cite{huang2017densely}, ResNeXt \cite{xie2017aggregated} to raise the performance bar on the benchmark datasets. 

Despite the massive success of CNNs on cross domains, there are several concerns with its mechanism of processing information. It is worthy to note that the \textit{neuronal design} of CNNs is still far away from that of the human visual system; this limits the CNNs from extracting and encoding human-level features from the data. Further, CNN's are inept at coping with the affine transformations as they cannot maintain spatial hierarchies amongst the features resulting in poor generalizability. Generally, a pooling operation is applied to reduce the spatial dimension of the feature space and thus to curtail the computation complexity of the network. Besides, pooling operation also leverages invariance and the ability to withstand a small amount of shift. However, the pooling operation comes at the cost of losing precise location and positional information, affecting CNN's ability to represent the actual distribution of the entity. Adding to it, as contrary to the mere functionality of the human brain, CNNs fail miserably in encoding the information from the small data. Furthermore, it is experimentally proven that the patterns encoded by CNNs are vulnerable and prone to adversarial attacks \cite{su2019one,moosavi2016deepfool,nguyen2015deep,liu2018intriguing}. From the above-discussed limitations, we can conclude that there is a need to promote analogous models working closely with the human visual system. 

In 2011, Hinton et al. \cite{hinton2011transforming} introduced the theory of \textit{capsules}. The capsules are the vectorized cluster of neurons, whose activity vector represents a specific type of entity's instantiation parameters, with the length of the vector depicting the probability of the presence of an entity. Inspired by Hinton's doctrine of capsules, Sabour et al. \cite{sabour2017dynamic} introduced a novel training mechanism called \textit{dynamic routing between the capsules} for iteratively training the capsule network architecture, which bestowed tremendous potential compared to the conventional CNNs on multiple datasets. The capsule network bypasses the sub-sampling operation by carefully employing the routing mechanism to propagate the elegant signals from lower-level layers to higher-level layers making the capsule network translational equivariance. The activated capsules in one layer make pose predictions for the capsules in the next layer via transformation matrices.  Then, the routing algorithm finds a center-of-mass of all the predictions using an iterative clustering algorithm. This approach ensures the propagation of relevant features to the subsequent layers, i.e., as more prediction vectors agree to the parent capsule, it gets a more significant activation. Besides, the capsule network can learn and encode the information from limited samples, making it exemplary from the conventional CNN models. Although capsule network possesses all the inclinations towards setting up a new standard in the neural networks domain, it comes with some inherent limitations \cite{sabour2017dynamic} such as 1) Capsule network incurs a large number of trainable parameters causing significant memory and hardware overhead. 2) The performance of the capsule network models is below par compared to the conventional CNN models on complex datasets/images involving complex background and foreground regions. 3) The capsule network tries to encode all the information from the cluttered or complex data, causing a detrimental effect in forming informative capsules.
In this work, we propose a novel capsule network-based architecture for efficiently dealing with complex images for image classification. The major contributions can be summarized as follows:
\begin{enumerate}
\item This paper presents a new approach for designing a capsule network architecture that utilizes a wide bottleneck residual connection, along with the SE blocks and attention capsules for accurately classifying complex images.
\item We extensively evaluated the performance of the proposed model on the three publicly available datasets, namely Fashion MNIST \cite{xiao2017fashion}, SVHN \cite{netzer2011reading}, and CIFAR-10 \cite{krizhevsky2009learning}, to achieve state-of-the-art performance on CIFAR-10 and Fashion MNIST with highly competitive performance on the SVHN dataset.
\end{enumerate}
The remainder of the paper is structured as follows: Section \ref{sec_one} presents an overview of the recent studies that made significant contributions in the capsule network literature of image classification. Section \ref{sec_two} provides a comprehensive overview of the proposed methodology. 
Section \ref{sec_three} narrates the data and experimental results. Section \ref{sec_four} concludes and provides a direction for future developments.
\section{Literature Survey}
\label{sec_one}
In this section, we outline the summary of various capsule network-based classification methods proposed in the literature. We classify these methods into 1) Structural modifications and 2) Algorithmic modification, based on the essence of the problem-solving approach. In Table \ref{tab1}, we have listed the performance analysis of various methods against the benchmark datasets.
\subsection{ Structural Modifications to Capsule Network}
Many methods have been proposed in the literature by adopting structural modification to improve the overall performance of the capsule network architectures.  Xiang et al. \cite{xiang2018ms} introduced a multi-scale capsule network or MS-CapsNet to comprehend the hardware overhead and to improve the feature representation capability. This method employs a multi-scale feature extraction block to achieve the objective. Further, they have introduced a \textit{capsule dropout} mechanism to introduce regularization in the network. MS-CapsNet achieved an accuracy of 92.70\%, 75.70\% on Fashion MNIST \cite{xiao2017fashion} and CIFAR-10 \cite{krizhevsky2009learning} datasets, respectively. Phaye et al. \cite{phaye2018dense} introduced two architectures, namely DCNet and DCNet++ to extract more complex features and thus to form efficient and robust primary capsules. DCNet incorporates DenseNet \cite{huang2017densely} architecture as the backbone to extrapolate complex and diversified features. DCNet achieved an accuracy of 99.75\% on the MNIST dataset.  DCNet++ incorporates the primary capsules that carry scale information at multiple levels to diversify the patterns encoded by the capsules to improve the performance on complex datasets. However, DCNet++ achieved a trivial improvement of 0.31\% over CIFAR 10 \cite{krizhevsky2009learning} dataset. 

Hoogi et al. \cite{hoogi2019self} proposed an attention-based capsule network called self-attention capsule network or SACN for accurately locating the region of interest or RoI from the complex images. Further, SACN is also capable of suppressing irrelevant and noisy features and thus improving the performance of capsule network architectures. SACN achieved reasonably good performance on SVHN \cite{netzer2011reading} and MNIST \cite{lecun1998mnist} datasets. Jia and Huang \cite{jia2020capsnet} proposed a diverse and enhanced capsule network or DE-CapsNet, which utilizes residual blocks and the position-wise dot product operation to develop enhanced primary capsules that carry scale information at multiple levels for dealing with complex data. Further, they adopted the softmax activation function over sigmoid in the dynamic routing algorithm for a better distribution of routing coefficients.
DE-CapsNet attained an accuracy of 94.25\%, 92.96\% on Fashion MNIST \cite{xiao2017fashion} and CIFAR-10 \cite{krizhevsky2009learning} datasets, respectively. Huang et al. \cite{huang2020capsnet} proposed a dual attention capsule network named DA-CapsNet to enable \textit{attention} in the capsule network. They incorporated the attention module in the convolution layer and primary capsule layer to form \textit{conv-attention} and \textit{caps-attention} to extract pertinent information. DA-CapsNet achieved an accuracy of 85.47\%, 94.82\%, 93.98\% on CIFAR-10 \cite{krizhevsky2009learning}, Fashion MNIST \cite{xiao2017fashion}, and SVHN \cite{netzer2011reading} datasets, respectively. Sun et al. \cite{sun2021novel} introduced dense capsule networks called DenseCaps to improve the performance of capsules on complex datasets. DenseCaps allows feature-resue and incorporates cross-capsule feature concatenation techniques to extract and encode salient features from complex data. DenseCaps achieved an accuracy of 99.70\%, 94.93\%, 95.99\%, and 89.41\% on MNIST \cite{lecun1998mnist}, Fashion MNIST \cite{xiao2017fashion}, SVHN \cite{netzer2011reading}, and CIFAR 10 datasets \cite{krizhevsky2009learning}, respectively.
\begin{table*}
\centering
\renewcommand{\arraystretch}{1.1}
\caption{ The Performance Comparison of Capsule Network based Methods with Benchmark Public Datasets.}
\begin{tabular}{@{}p{4.0cm}cccccccccc@{}}
\toprule
{Methods} &  \multicolumn{4}{c}{Accuracy} \\
\cmidrule(lr){2-5}
& MNIST \cite{lecun1998mnist} & 
SVHN \cite{netzer2011reading} & F-MNIST \cite{xiao2017fashion} & CIFAR-10 \cite{krizhevsky2009learning}\\ \midrule
Deliege et al. \cite{deliege2018hitnet} & 99.62 & 94.50 & 92.30 & 73.30 \\

Xiang et al. \cite{xiang2018ms} & -- &  -- & 92.70 & 75.70 \\

Huang et al. \cite{huang2020capsnet} & -- & 93.98 & 94.82 & 85.47 \\

Hinton et al.\cite{hinton2018matrix}& --  & -- & -- & 88.01\\

Sun et al.\cite{sun2021novel}& 99.70  & 95.99 & 94.93 & 89.41\\

Phaye et al. \cite{phaye2018dense} & 99.71 & 95.58 & 94.65 & 89.71\\

Rajasegaran et al. \cite{rajasegaran2019deepcaps} & -- &  -- & 94.73 & 92.74   \\

Jia and Huang \cite{jia2020capsnet}& --  & -- & 94.25 & 92.96 \\

Zhao et al. \cite{zhao2019capsule} & -- & -- & 94.70 & 93.20 \\

Yang et al. \cite{yang2020rs} & -- & 97.08 & -- & 93.32 \\

Fuchs and Pernkopf \cite{fuchs2020wasserstein} & 99.68 & 96.56 &-- & 93.43 \\

Rezwan et al. \cite{rezwan2020mixcaps} & -- & -- &-- & 94.72 \\

Sun et al. \cite{sun2020deep} & -- & 94.41 & 95.50 & -- \\

Tsai et al. \cite{tsai2020capsules} & -- & -- &-- & 95.14 \\
\bottomrule
\end{tabular}
\label{tab1}
\end{table*}

\subsection{ Algorithmic Modifications to Capsule Network}
To mitigate the computation complexity and the massive amount of transformation matrices incurred by Sabour's method \cite{sabour2017dynamic}, Hinton et al. \cite{hinton2018matrix} introduced a new routing method called expectation maximization or EM routing algorithm. Further, they have proposed \textit{matrix capsules} to encode the properties of an entity or feature. Capsules trained with the EM algorithm attained state-of-the-art results on the SmallNorb dataset \cite{lecun2004learning}; however, its performance on CIFAR-10 \cite{krizhevsky2009learning} remained sub-par. Deliege et al. \cite{deliege2018hitnet} introduced HitNet that employs a novel hit or miss (HoM) layer with the centripetal loss function by substituting the classification capsules and the margin loss function of Sabour's method \cite{sabour2017dynamic}. Also, the authors have introduced \textit{ghost capsules} for classifying the mislabelled training data. HitNet achieved an accuracy of 83.03\%, 94.50\%, 73.30\%, 92.30\%, 99.62\% on affNIST \cite{affNIST}, SVHN \cite{netzer2011reading}, CIFAR-10 \cite{krizhevsky2009learning}, Fashion MNIST \cite{xiao2017fashion}, and MNIST \cite{lecun1998mnist} datasets, respectively.
Wang et al. \cite{wang2018optimization} addressed the routing algorithm as an optimization problem and introduced a novel routing technique similar to the agglomerative fuzzy k-means algorithm. They have experimentally demonstrated the efficacy of their method in comparison with Sabour's method \cite{sabour2017dynamic}. Fuchs and Pernkopf \cite{fuchs2020wasserstein} introduced the Wasserstein objective function to optimize the performance of the capsule network. The Wasserstein capsule network or W-CapsNet facilitated scalability with relatively fewer trainable parameters to improve the performance on the complex datasets. W-CapsNet achieved an accuracy of 70.39\%, 93.43\% on the CIFAR-100 and CIFAR-10 \cite{krizhevsky2009learning} datasets, respectively. 

To subdue the expensive computation complexity and the requirement of powerful resources, Zhao et al. \cite{zhao2020efficient} proposed factorized machines (FM) routing algorithm. Unlike Sabour's approach, where all the primary capsules contributes in predicting the output of the secondary-level capsules; in FM routing, the pair capsules will look for an agreement among themselves; if there is a strong agreement, these capsules approximate the secondary capsules output. The authors have evaluated the performance of their model on affNIST \cite{affNIST}, SVHN \cite{netzer2011reading}, CIFAR-10 \cite{krizhevsky2009learning}, and MNIST \cite{lecun1998mnist} datasets, and achieved an accuracy of 93.85\%, 96.79\%, 93.70\%, and 94.70\%, respectively. Rezwan et al. \cite{rezwan2020mixcaps} introduced a simplistic data routing technique called iteration-free rating to minimize the number of routing iterations required for the convergence of capsule network architecture. Further, the method in \cite{rezwan2020mixcaps} method focuses on curtailing the training time and resource overhead.
This architecture makes use of a new class of capsules called a \textit{single matrix} to achieve the objective. Iteration-free routing attained an accuracy of 75.85\%, 94.72\% on CIFAR-100 and CIFAR-10 \cite{krizhevsky2009learning} datasets. 

Rajasegaran et al. \cite{rajasegaran2019deepcaps} proposed a new approach called DeepCaps by introducing a 3D convolution-based routing algorithm to expedite depth in the capsule network architecture by curtailing the number of trainable parameters. They have also introduced a class independent decoder as a regularizer to aid the performance. DeepCaps attained an accuracy of 99.72\%, 94.73\%, 97.56\%, 92.74\% on MNIST \cite{lecun1998mnist}, Fashion MNIST \cite{xiao2017fashion}, SVHN \cite{netzer2011reading}, and CIFAR-10 \cite{krizhevsky2009learning} datasets, respectively. Tsai et al. \cite{tsai2020capsules} introduced a new routing algorithm by employing the following design changes, 1) layer normalization, 2) adopting concurrent routing instead of iterative routing, and 3) routing with inverted dot product attention. This method achieved an accuracy of 95.14\%, 78.02\% on CIFAR 10 and CIFAR-100 \cite{krizhevsky2009learning} datasets, respectively.
A tensor-based capsule network called DeepTensorCaps was introduced by Sun et al. \cite{sun2020deep} to uplift the performance of capsule networks on complex data. Unlike the vector capsules, they have claimed that the tensor capsules are assertive in encoding the complex features. DeepTensorCaps incorporates \textit{tensor dropout} and multi-scale decoder networks as the regularizers to aid the performance of the model. DeepTensorCaps archived an accuracy of 97.41\%, 92.87\%, 95.50\% on SVHN \cite{netzer2011reading}, CIFAR 10 \cite{krizhevsky2009learning}, and Fashion MNIST \cite{xiao2017fashion} datasets, respectively. Motivated by Sabour's approach \cite{sabour2017dynamic} of representing the features as \textit{parse-tree} structure, Yang et al. \cite{yang2020rs} introduced RS-CapsNet to form \textit{bigger-part} with the help of intermediate-capsules to improve the overall performance of the capsule network. RS-CapsNet achieved an accuracy of 97.08\%, 91.32\%, 94.08\% on SVHN \cite{netzer2011reading}, CIFAR10 \cite{krizhevsky2009learning}, and Fashion MNIST \cite{xiao2017fashion} datasets, respectively.
\section{Proposed Methodology}
\label{sec_two}
\begin{figure*}
\centering
\includegraphics[scale=0.050]{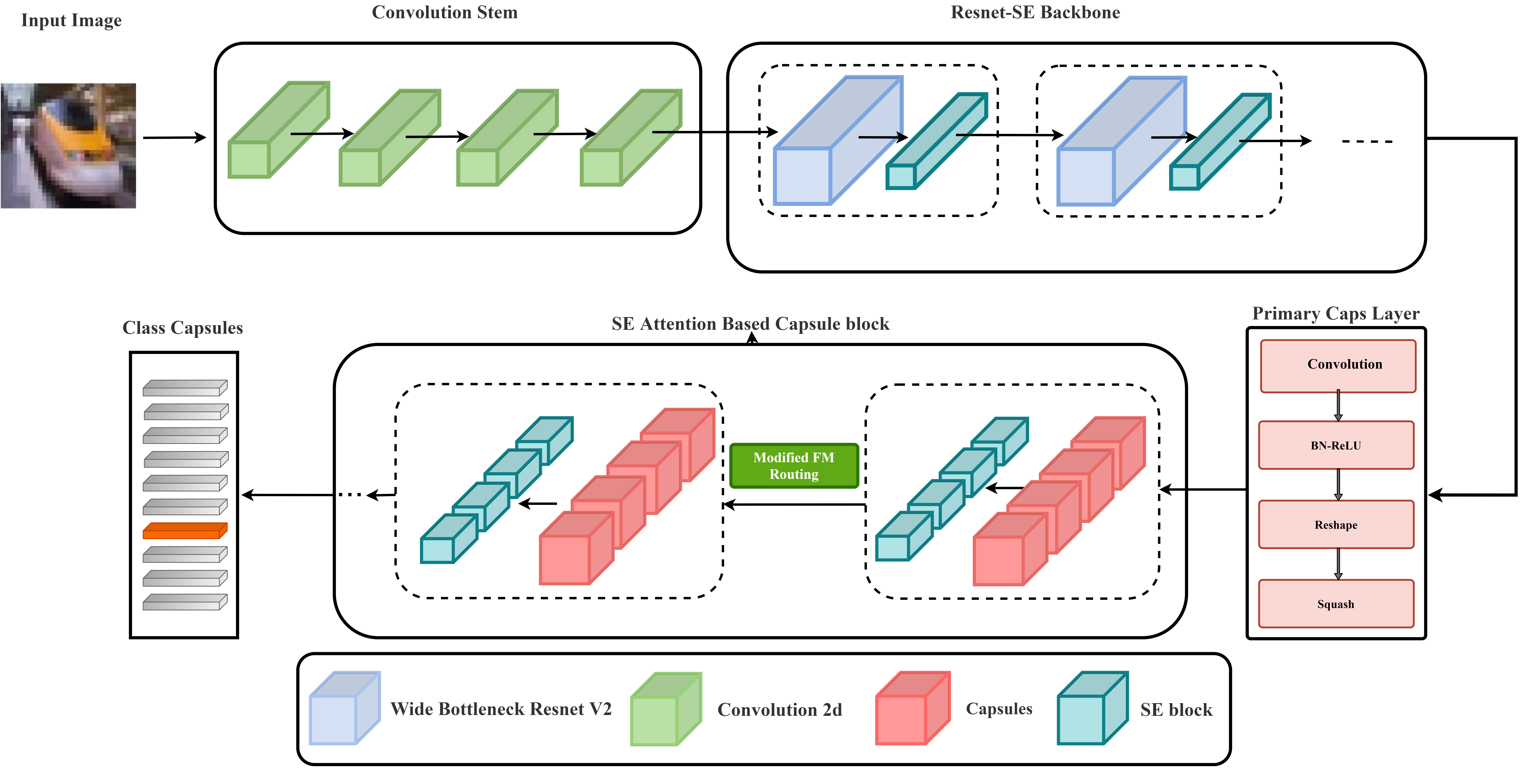}
\caption{An Overview of the Proposed Wide Capsule Network Architecture.}
\label{dia1}
\end{figure*}

Having discerned the limitations of the capsule network architectures proposed in the literature, the research in capsule networks for image classification headed towards fusing CNN with the capsules by retaining the \textit{capsulness} nature in the network. In this direction, we introduce a wide capsule network or WideCaps to address the inherent limitations of the capsule network architectures on complex datasets. Fig. \ref{dia1} depicts the block diagram of the proposed WideCaps architecture. We adopt a wide bottleneck residual module along with SE-block as the backbone network to form informative primary capsules. This follows the attention capsule that suppresses the noise and irrelevant features by allowing the propagation of salient features to higher-level capsules. Also, we propose a modified FM routing technique that guarantees the appropriate distribution of routing coefficients and thus building an efficient workflow for capsule network-based image classification.
\subsection{The Wide Bottleneck Residual and SE block}
Inspired by the recent success of normalizer-free networks \cite{brock2021high}, we strive to achieve emphatic primary capsules and thus configuring an efficient classification workflow. A wide capsule network architecture entails a convolutional stem followed by the wide bottleneck residual and SE block as the backbone architecture. An input image is passed onto the stem of 4 successive convolutional layers, and the resultant output is subjected to the subsequent backbone architecture to generate salient feature maps. The convolutional stem comprises four 2D convolutional layers with 16, 32, 64, 128 kernels each of shape $(3\times 3)$ having strides $(1\times1)$ with an interleaved  ReLU \cite{maas2013rectifier} activation function. The output of the convolutional stem is then passed onto the backbone architecture to generate the salient feature maps. The backbone architecture contains three successive wide bottleneck residual and SE blocks in sequence. Each block has 4, 8, and 4 ResNet-SE layers, respectively. 

\begin{figure*}
\centering
\includegraphics[scale=0.10]{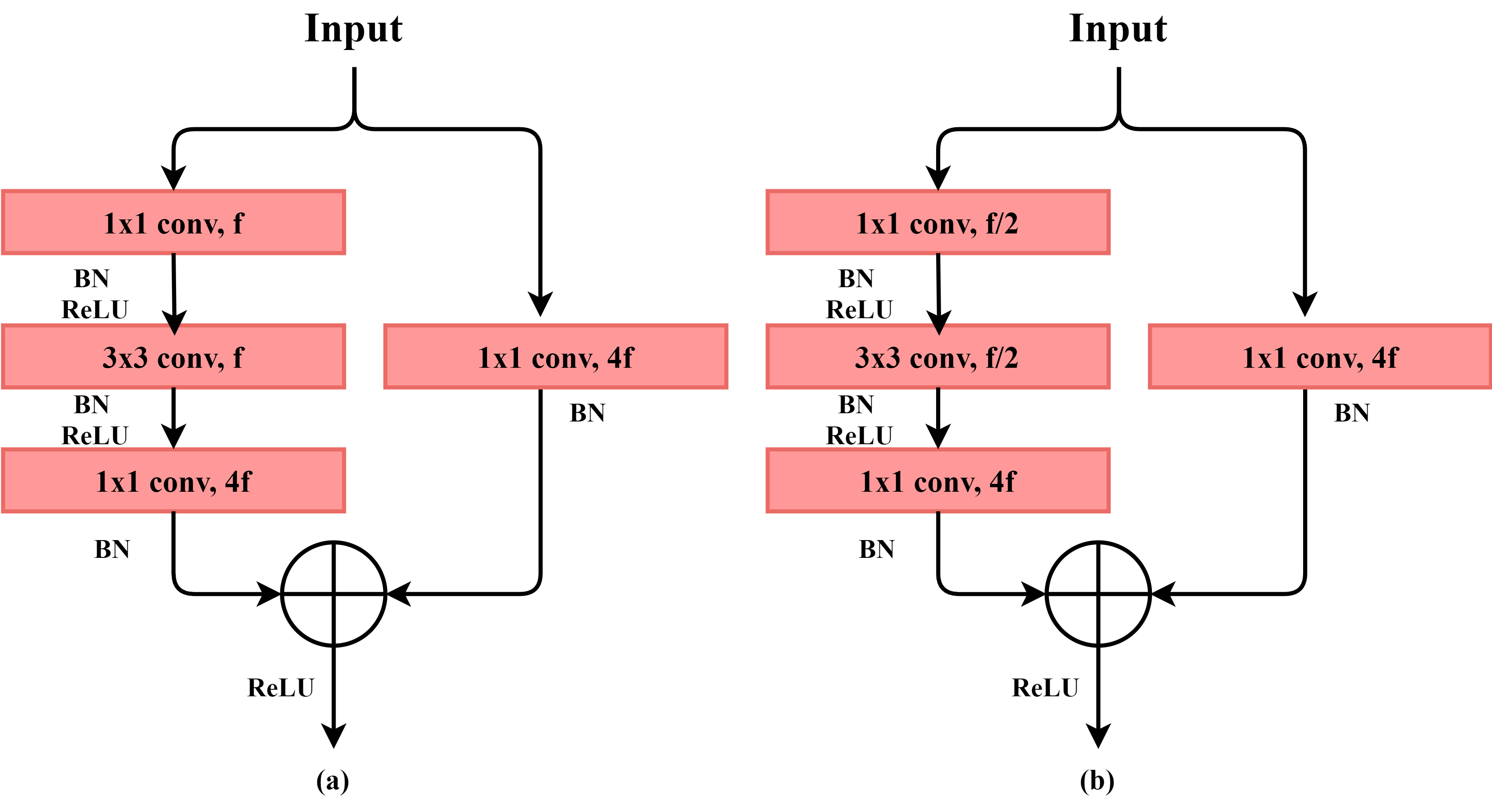}
\caption{A Block Diagram of (a) Standard Bottleneck Residual Block and (b) Wide Bottleneck Residual Block.}
\label{dia10}
\end{figure*}

We adopt a wide bottleneck residual block (all the bottleneck residual blocks are of projection version) over the standard bottleneck residual block; due to its efficient performance with fewer kernels, thereby curtailing the computation cost without affecting the performance. Fig. \ref{dia10} represents the schematic diagram of a standard bottleneck residual block (a) and wide bottleneck residual block (b). The bottleneck residual block includes three convolutional layers with filters $(1\times1)$, $(3\times3)$, and $(1\times1)$, respectively interleaved with ReLU activation and Batch Normalization. Additionally, the block incorporates a $(1\times1)$ convolutional layer in the skip connections with an appropriate number of filters to match the dimension. Finally, the feature maps from both ends are summed up, and the resultant output is passed onto the ReLU activation, followed by the rest of the network. We modify the bottleneck residual block by halving the number of filters $f$ of the bottleneck residual block to $f/2$ in the first two convolutional layers, followed by multiplying it with 4 to regain the original number of filters to form a wide bottleneck residual block. Integrating the wide bottleneck residual block  with SE blocks helps in formulating robust and competent feature maps. The output of the backbone architecture is reshaped to form primary capsules by the primary capsule layer. Inside the primary capsule layer, a convolutional layer downsamples the tensor generated by the backbone architecture, followed by the batch normalization layer. This tensor is further converted into capsules by reshaping (e.g., a tensor with shape $H$ × $W$ × $C$ is reshaped into $H$ × $W$ × $C\textsuperscript{'}$ × $K$; where, $H$, $W$, $C$ denotes height, width, channel respectively, and $K$ is the number of neurons in a capsule), followed by a non-linear \textit{squash}  activation function \cite{sabour2017dynamic}. Further, the output capsules of the primary capsules are passed onto the attention-based capsules to generate output capsules. 
\begin{figure*}
\centering
\includegraphics[scale=0.048]{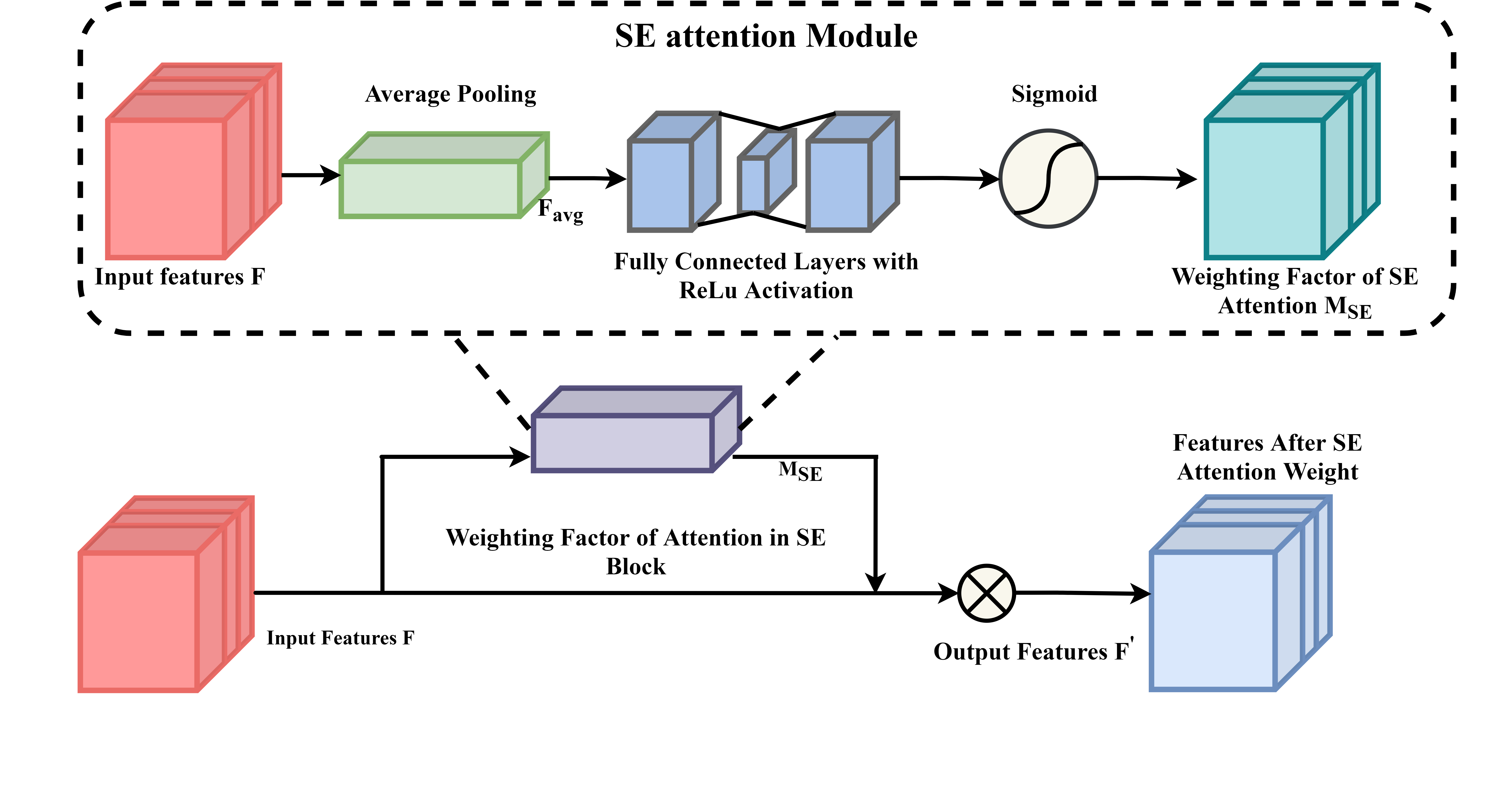}
\caption{ Squeeze and Excitation Attention Block. }
\label{dia2}
\end{figure*}
\subsection{Squeeze and Excitation Attention based Capsules}
The capsule network has an inherent limitation of being responsive to noisy and irrelevant features causing significant performance degradation in complex images. To alleviate this, we introduced attention-based capsules that focus on the most prominent parts of the image by diminishing the features that are unlikely to have any significance. Fig. \ref{dia2} represents the block diagram of SE-attention-based capsules. The attention mechanism guides the capsules \textit{where to look} for essential  features in the image by directing the attention of the network using depth features captured by the capsules. The attention mechanism uses this correlation extracted by the capsules to generate attention maps. Further, it helps to improve the channel inter-dependencies, thereby enhancing the overall representation ability of the capsules.
\begin{table*}\centering
\renewcommand{\arraystretch}{1.0}
\hspace{1.5em}
\caption{ The Architectural Design of the Proposed WideCaps Method.}
\begin{tabular}{@{}p{1.9cm}cccccccccc@{}}\toprule
{Block} & {Type} & {Operation}\\
\midrule
Stem & Convolution & Conv, [3$\times$3]$\times$16\\
&  & Conv, [3$\times$3] $\times$ 32\\
&  & Conv, [3$\times$3] $\times$ 64\\
&  & Conv, [3$\times$3] $\times$ 128\\
\\
Backbone & Wide-Bottleneck Residual SE & Conv, [1$\times$1]$\times$16\\
&  & Conv, [3$\times$3]$\times$32\\
&  & Conv, [1$\times$1]$\times$64\\
&  &       fc,[16$\times$64]\\
\\
Capsules & Wide-Bottleneck Residual SE
&    Conv, [3 $\times$ 3] $\times$16 \\
&  & Conv, [3 $\times$ 3] $\times$ 128\\
&  & Conv, [3 $\times$ 3] $\times$ 128\\
&  & fc,[16 $\times$ 128] \\[2.5pt]
\bottomrule
\end{tabular}
\label{tab2}
\end{table*}

The attention capsules are formed by transforming the classification capsules into a vector represented by \textit{F}, as shown in Fig. \ref{dia2}. In the subsequent stage, the resultant vector \textit{F} is propagated to the SE attention module. In the SE module, the features are subjected to a squeeze operation, wherein every channel of the input feature \textit{F} is squeezed to a single numeric value by applying average pooling operation resulting in $F_{avg}$, and then passed to the excitation operation. The excitation block consists of 2 fully connected layers activated by ReLU \cite{maas2013rectifier} and sigmoid operation functions \cite{jamel2012implementation} to form the feature vectors $O_{ReLU}$ and $M_{SE}$, respectively. Finally, the output of the SE attention model is multiplied with input features $F$ to get the output features $F\textsuperscript{'}$. The Eqs \ref{eq1}-\ref{eq4} \cite{huang2020capsnet} delineates the working of SE-attention model. Table \ref{tab2} provides the architectural detail of the proposed method.
\begin{equation}
\mathrm{F_{avg}} = \mathrm{AvgPooling(F)}
\label{eq1}
\end{equation}
\begin{equation}
\mathrm{O_{ReLU}} =  \mathrm{ReLU(W_1 * F_{avg} + b_1)}    
\label{eq2}
\end{equation}
\begin{equation}
\mathrm{M_{SE}} =  \mathrm{sigmoid(W_2 * O_{ReLU} + b_2)}
\label{eq3}
\end{equation}
\begin{equation}
\mathrm{F\textsuperscript{'}} = \mathrm{F * M_{SE}}
\label{eq4}
\end{equation}
\subsection{Modified FM Routing Algorithm}
\label{algo}
We introduce a modified FM routing algorithm to keep the range of prediction vectors bounded and to have the routing coefficients well distributed to improve the correlation between the parent and child capsules. We achieve this by adopting the softmax activation function over \textit{exp} activation adopted by \cite{zhao2020efficient}, which resulted in the uneven distribution of routing coefficients; whereas, softmax enables the relevant features to get a high value and the counterpart. The working of the modified FM routing algorithm is presented below.

Every capsule in the layer $l-1$ is denoted by $u_i$, and the prediction vector $\hat{u}_{j|i}$ for every capsule in the layer $l$ is computed. Let the set of prediction vectors be [$\hat{u}_{j|0}$, $\hat{u}_{j|1}$.....$\hat{u}_{j|n}$], where the agreement between the capsules is formed through pairwise interactions amongst the capsules in the same layer. The pairwise product is established as, $\hat{b}_{j|i_1,i_2} = \hat{u}_{j|i_1} \odot \hat{u}_{j|i_2}$, where the summation of each element of $\hat{b}_{j|i_1,i_2}$ gives the magnitude of agreement and also the orientation and pose of capsule $j$. In general, the pairwise interaction of all capsules in layer $l-1$ to the capsule $j$ of layer $l$ can be formulated as in Eq \ref{eq5} \cite{zhao2020efficient},
\begin{equation}
\begin{split}
\mathrm{\hat{H_j}} & = \mathrm{\sum_{i_1 = 1}^{n} \sum_{i_2 = i_1 + 1}^{n} \hat{u}_{j|i_1} \odot \hat{u}_{j|i_2}}\\
& = \mathrm{\frac{1}{2} \left(\sum_{i=1}^{n}\hat{u}_{j|i} \odot \sum_{i=1}^{n}\hat{u}_{j|i} - \sum_{i=1}^{n}\hat{u}_{j|i} \odot \hat{u}_{j|i}\right)}
\label{eq5}
\end{split}
\end{equation}
where $\hat{u}_{j|i} = [\hat{u}_{j|1}, \hat{u}_{j|2}, ..... \hat{u}_{j|n}]$, $\hat{H}_{j} = [\hat{H}_{j,1}, \hat{H}_{j,1}, ..... \hat{H}_{j,n}]$, and $n$ denote the total number of prediction vectors. Then the output of capsule $j$ in layer $l$ is defined as in Eq \ref{eq6} \cite{zhao2020efficient},
\begin{equation}
\mathrm{\hat{b}_j} = \mathrm{\sum_{f=1}^{k} \hat{H}_{j,f}}
\label{eq6}
\end{equation}
The coefficients of agreement is calculated by applying softmax function on the output as depicted in Eq \ref{eq7}.
\begin{equation}
\mathrm{\hat{x}_{j}} =  \mathrm{softmax(\hat{b}_j)}
\label{eq7}
\end{equation}
The pose vector is defined as $\hat{Q}_j = \frac{\hat{H}_j}{\left \| \hat{H}_j \right \|}$, where the direction of $\hat{Q}$ determines the pose, orientation, size, rotation, etc of an entity. As the summation operations present in the Eqs.\ref{eq5}\ref{eq6}, could lead to the possible gradient explosion resulting in poor performance, the prediction vector $\hat{u}_{j|i}$ is scaled by  dividing it with $\sqrt{n}$  \cite{zhao2020efficient}, which results in Eq. \ref{eq8} \cite{zhao2020efficient}:
\begin{equation}
\mathrm{\hat{H}_j} = \mathrm{\frac{1}{2n} \left(\sum_{i=1}^{n}\hat{u}_{j|i} \odot \sum_{i=1}^{n}\hat{u}_{j|i} - \sum_{i=1}^{n}\hat{u}_{j|i} \odot \hat{u}_{j|i}\right)}
\label{eq8}
\end{equation}
\begin{algorithm}[H]
\renewcommand{\arraystretch}{1.4}
\caption{Modified FM Routing}
 \hspace*{\algorithmicindent} \textbf{Input: } {Prediction vectors $\hat{u}_j = (\hat{u}_{j|1}, \hat{u}_{j|2}, ...., \hat{u}_{j|n})$} \\
 \hspace*{\algorithmicindent} \textbf{Output: } {$\hat{Q}_j, \hat{x}_j$}
\begin{algorithmic}[1]
\STATE  $\mathbf{\hat{u}_{j|i}}\leftarrow L2Normalize(\hat{u}_{j|i})$ 
\STATE $\mathbf{\hat{H}_j}\leftarrow \frac{1}{2n} (\sum_{i=1}^{n}\hat{u}_{j|i} \odot \sum_{i=1}^{n}\hat{u}_{j|i} - \sum_{i=1}^{n}\hat{u}_{j|i} \odot \hat{u}_{j|i})$ 
\STATE $\mathbf{\hat{Q}_j}\leftarrow\frac{\hat{H}_j}{\left \| \hat{H}_j \right \|}$
\STATE $\mathbf{\hat{b}_j}\leftarrow \sum_{f=1}^{k} \hat{H}_{j,f}$ 
\STATE $\mathbf{\hat{x}_{j}}\leftarrow softmax(\hat{b}_j)$ 
\end{algorithmic}
\label{one}
\end{algorithm}
We use $\hat{x}_j$ to determine which class is activated and by how much it has been activated.
Combing all the declarations and equations defined above, we conclude the above process in Algorithm \ref{one}, to be the modified version of FM routing \cite{zhao2020efficient}.
\begin{figure*}
\centering
\includegraphics[scale=0.12]{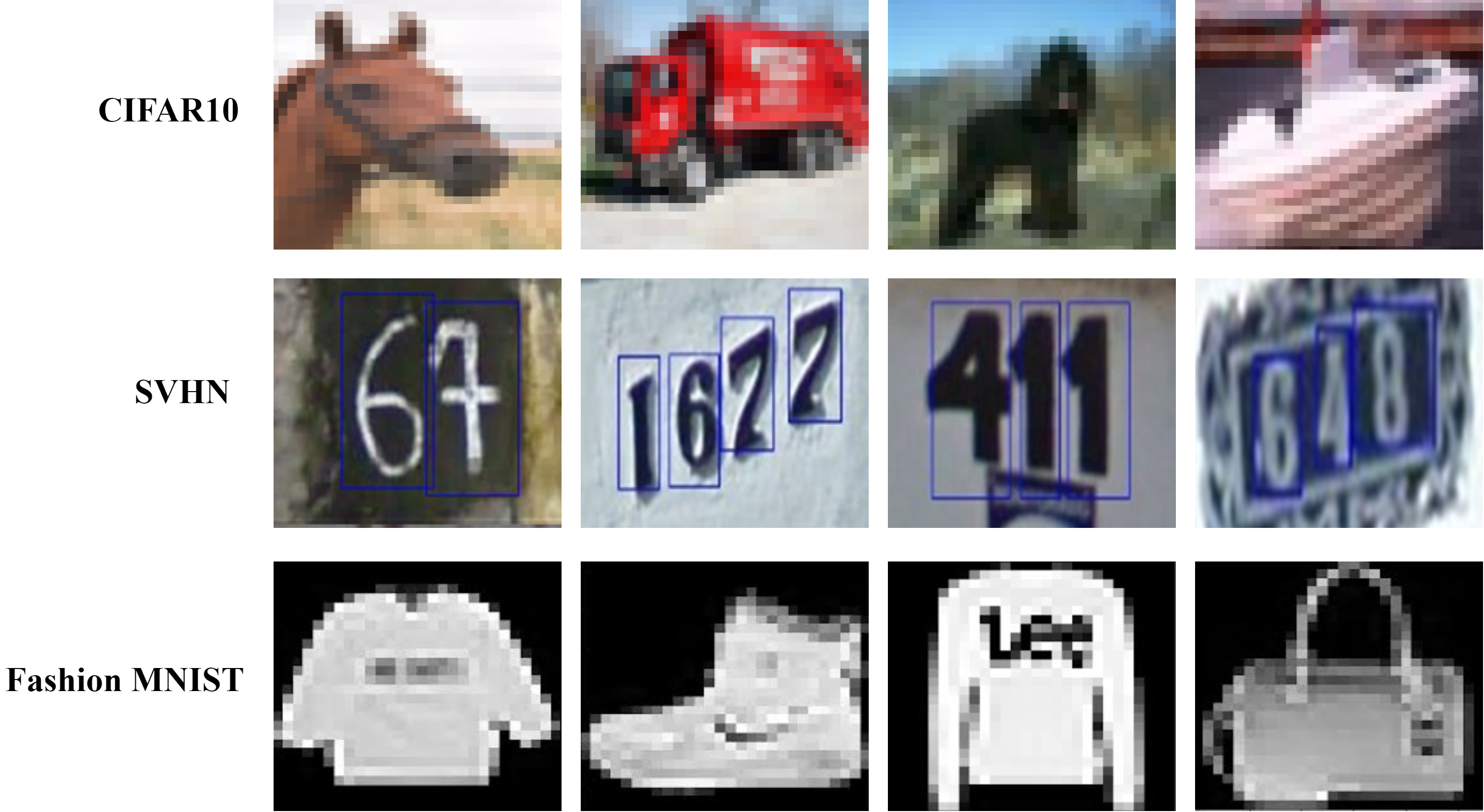}
\caption{Sample Images from the Datasets Adopted in the Study.}
\label{dia5}
\end{figure*}
\section{Datasets and Experimental Results}
\label{sec_three}
In this section, we brief on the datasets adopted in the study to evaluate the performance of the proposed model, followed by the hardware setup and training methodology. Further, we also compare and present the quantitative performance of the proposed method with state-of-the-art methods.  
\subsection{Dataset Description}
\begin{enumerate}
    \item Fashion MNIST dataset \cite{xiao2017fashion}:
    The Fashion MNIST dataset consists of 70,000 grayscale images, with 60,000  images for training and 10,000  images for testing. The dataset is a collection of 10 fashion items, with 7000 samples per class. The dataset was introduced as a substitute to the MNIST dataset \cite{lecun1998mnist} with slightly increased complexity level and is extensively used to evaluate the performance of capsule network models due to its higher complexity level than the naive MNIST dataset \cite{lecun1998mnist}.
    \item Street View House Numbers dataset \cite{netzer2011reading}:
    SVHN is a real-time RGB dataset. It contains house numbers collected from Google Street View Images. SVHN is widely adopted for evaluating the performance analysis of machine learning models.  The dataset consists of 70,000 images for training and 30,000 images for testing. The dataset includes images with diverse backgrounds, and hence it is widely used to evaluate the performance of capsule network-based models.
    \item Canadian Institute for Advanced Research-10 dataset \cite{krizhevsky2009learning}: CIFAR-10 is the most widely used tiny RGB dataset of 10 distinct classes acquired with a diverse background and intensity level. The dataset consists of 60,000 images for training and 10,000 images for testing. CIFAR-10 is the most complex data in comparison with the Fashion MNIST \cite{xiao2017fashion}, and SVHN \cite{netzer2011reading}, as the performance of capsule network-based architectures is sub-par on complex data, CIFAR-10 is extensively used to evaluate the performance of capsule network-based architectures. Fig. \ref{dia5} shows the sample images from the above-discussed datasets.
\end{enumerate}
\subsection{Hardware Setup and Training Methodology}
All the experiments were conducted on a DGX-1 workstation having 8$\times$NVIDIA Tesla V100 GPUs with 256 GB dedicated memory on a Ubuntu 18.04 operating system, 64-bit Intel Xeon(R) Gold 5120 CPU @2.20 GHz × 28 processor, solid-state hard drive, 64 GB RAM. Further, the implementation was done in Python 3.6 with Keras chollet2015keras and TensorFlow 2 \cite{tensorflow2015-whitepaper} as backend.   

The initial weight matrix is initialized with \textit{HeNormal} distribution \cite{he2015delving}. We adopted an L2 regularizer to regulate the weight matrix. Further, we used Stochastic Gradient Descent with momentum as the optimizer to train the network with a decay rate of 0.5 and momentum of 0.9. We adopted a drop-based learning rate scheduler strategy as shown in Eq \ref{ten} \cite{chollet2015keras}, with an initial learning rate (ILR) of 0.01, that halves the learning rate (LR) at every fixed number of epochs during the training process. In our case, we empirically found dropping the learning rate at every 60 epochs would give better performance.
\begin{equation}
    \mathrm{LR = ILR * DropRate\textsuperscript{(Epoch / Epoch Drop)}}
    \label{ten}
\end{equation}
Here, DropRate is the rate at which the learning rate to be changed, and EpochDrop is how frequently to change the learning rate. We train our model from scratch with a  batch size of 128 till the convergence with softmax categorical cross-entropy loss function \cite{bruch2019analysis} to better estimate the performance.
\subsection{Results and Discussion}
\begin{figure*}[h!]
\centering
\includegraphics[width=16cm]{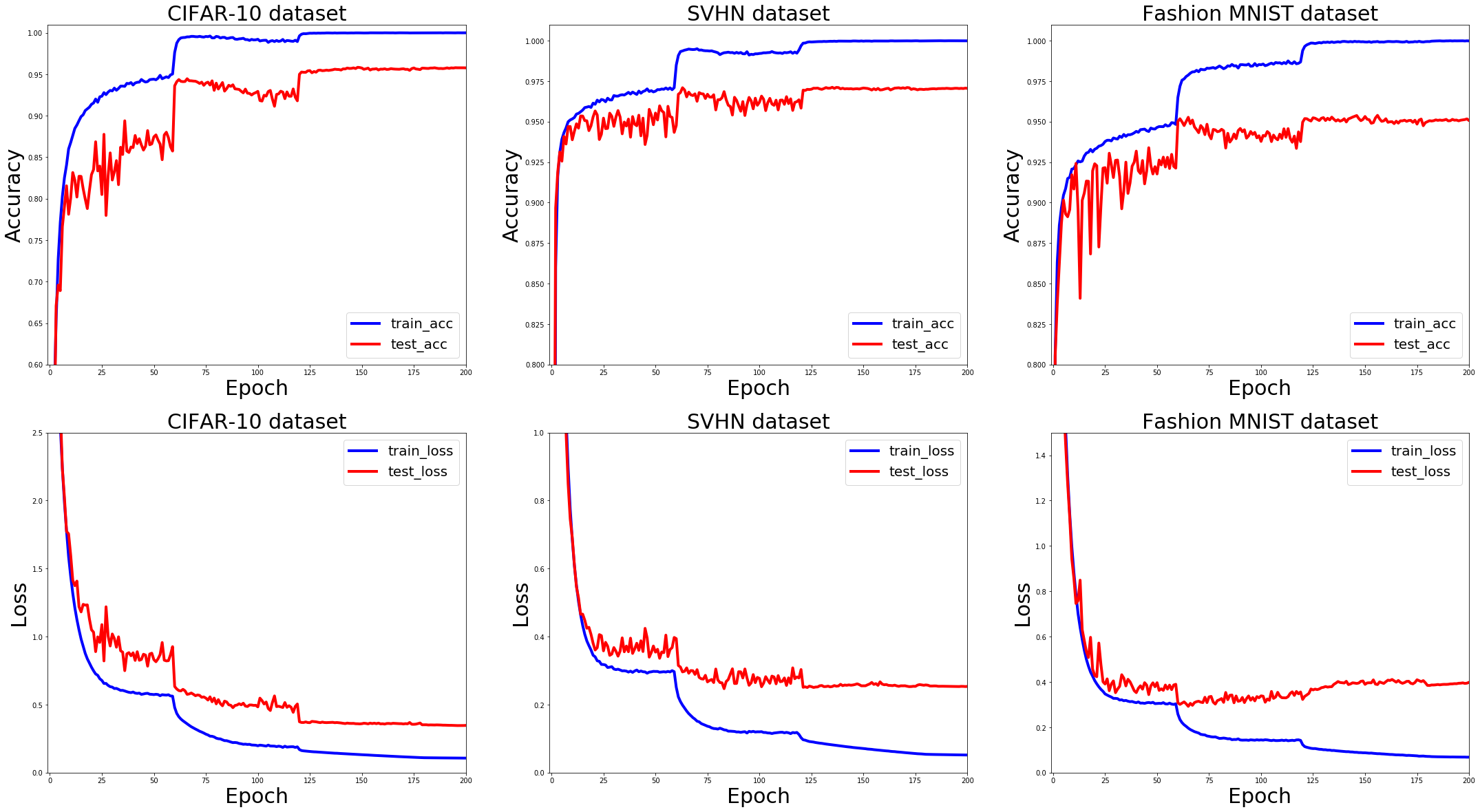}
\caption{Learning Curve of the Proposed Method on the Benchmark Datasets.}
\label{dia3}
\end{figure*}

We quantitatively evaluate and compare the performance of the proposed method with the other methods proposed in the literature by considering accuracy as the metric as defined in Eq \ref{eq9}. In Table \ref{tab3}, we compare the efficacy of the proposed model with top-5 accuracy on Fashion MNIST \cite{xiao2017fashion}, SVHN \cite{netzer2011reading}, and CIFAR-10 \cite{krizhevsky2009learning} datasets, respectively. It is apparent from Table \ref{tab3} that the proposed method outperforms the top-5 accuracy on Fashion MNIST \cite{xiao2017fashion} and CIFAR-10 \cite{krizhevsky2009learning} datasets  with highly competitive performance on the SVHN dataset \cite{netzer2011reading}. Fig. \ref{dia3} represents the learning curve across the three benchmark datasets. 
\begin{table*}
\renewcommand{\arraystretch}{1.3}
\centering
\hspace{1.5cm}
\caption{ The Performance Comparison of the Proposed Method with top-5 Accuracy on Fashion MNIST \cite{xiao2017fashion}, SVHN \cite{netzer2011reading}, and CIFAR-10 \cite{krizhevsky2009learning} Datasets, respectively.}
\begin{tabular}{@{}p{0.65cm}cccccccccc@{}}\toprule
Dataset & 
\multicolumn{2}{c}{Fashion MNIST\cite{xiao2017fashion}} &
\multicolumn{2}{c}{SVHN \cite{netzer2011reading}} &
\multicolumn{2}{c}{CIFAR-10 \cite{krizhevsky2009learning}} \\
\cmidrule(lr){2-3} \cmidrule(lr){4-5} \cmidrule(lr){6-7}
& Methods & Accuracy & Methods & Accuracy & Methods & Accuracy \\ \midrule
1 & Sun et al. \cite{sun2020deep} & 95.50  & Sun et al. \cite{sun2020deep} & \textbf{97.41} & Tsai et al. \cite{tsai2020capsules} & 95.14 \\
2 & Rajasegaran et al. \cite{rajasegaran2019deepcaps} & 94.73  & Yang et al. \cite{yang2020rs} & 97.08 & Rezwan et al. \cite{rezwan2020mixcaps} & 94.72 \\
3 & Zhao et al. \cite{zhao2020efficient} & 94.70  & Fuchs et al. \cite{fuchs2020wasserstein} & 96.56 & Fuchs et al. \cite{fuchs2020wasserstein} & 93.43 \\
4 & Phaye et al. \cite{phaye2018dense} & 94.65  & Phaye et al. \cite{phaye2018dense} & 95.58 & Yang et al. \cite{yang2020rs} & 93.32 \\
5 & Sabour et al. \cite{sabour2017dynamic} & 94.65  & -- & -- & Zhao et al. \cite{zhao2019capsule} & 93.20 \\
& Proposed & \textbf{95.59}  &   & 97.26 &  & \textbf{96.01} \\
\bottomrule
\end{tabular}
\\[5pt]
\label{tab3}
\end{table*}

\begin{table*}\centering
\renewcommand{\arraystretch}{1.3}
\caption{The Performance Comparison of Different Versions of the Proposed Method on the CIFAR-10 \cite{krizhevsky2009learning} Dataset, PC: Primary Capsule,  CC: Classification Capsule, Modified FM-R: Modified FM-Routing}
\begin{tabular}{@{}p{3.9cm}cccccccccc@{}}\toprule
\textbf{Methods} & \textbf{Modifications} & \textbf{Accuracy} \\ \midrule
Sabour et al. \cite{sabour2017dynamic} &Conv$\rightarrow$PC$\rightarrow$Dynamic Routing$\rightarrow$CC & 89.71\\
Zhao et al. \cite{zhao2020efficient} & ResNetv2$\rightarrow$PC$\rightarrow$FM-R$\rightarrow$CC & 93.20 \\
WideCaps v1 &Wide ResNet$\rightarrow$PC$\rightarrow$FM-R$\rightarrow$CC & 94.66 \\
WideCaps v2 &Wide ResNet$\rightarrow$PC$\rightarrow$Modified FM-R$\rightarrow$CC & 95.15 \\
WideCaps v3 &Wide ResNet+SE$\rightarrow$PC$\rightarrow$Modified FM-R$\rightarrow$CC & 95.35 \\
WideCaps v4 &Wide BottleNeck ResNet+SE$\rightarrow$PC$\rightarrow$Modified FM-R$\rightarrow$CC & 95.71 \\
WideCaps v5/\textbf{Proposed} &Wide BottleNeck ResNet+SE $\rightarrow$PC $\rightarrow$Modified FM-R $\rightarrow$ & \\
&Attention-based CC & \textbf{96.01} \\
\bottomrule
\end{tabular}
\label{tab4}
\end{table*}

\begin{equation}
    \mathrm{Accuracy = \frac{TP + TN}{TP + TN + FP + FN}}
\label{eq9}
\end{equation}
The study centered on improving the performance of the proposed capsule network-based model on the CIFAR 10 \cite{krizhevsky2009learning} dataset as it involves numerous challenges such as complex features, distinct backgrounds with various noise levels. Also, it's more likely that if a model could perform well on complex datasets, then we can expect reasonably good performance on less complex datasets such as SVHN \cite{netzer2011reading} and Fashion MNIST \cite{xiao2017fashion} datasets. 
So, all the structural, algorithmic modifications followed by the hyperparameter tuning were based on the results obtained for the CIFAR-10 dataset. \cite{krizhevsky2009learning}.  We can observe from Table \ref{tab3} that the proposed model achieved 6.39 \% improvement over the benchmark model \cite{sabour2017dynamic} and 0.8\% improvement over the state-of-the-art, which we believe is significant. However, it's worth noting that the result obtained for the benchmark model was based on the ensemble of 7 models. Adding to it, the proposed method improved the performance on the Fashion MNIST dataset \cite{xiao2017fashion} by 0.10\% on the state-of-the-art with highly competitive performance over the SVHN dataset \cite{netzer2011reading}. 

As we discussed in the previous section, the benchmark method \cite{sabour2017dynamic} was using a naive setup for forming the primary capsules and propagating the information downstream, which became a primary cause for the performance degradation on complex datasets. To tackle this, many researchers adopted powerful CNN-based architectures as the backend to form more robust and informative capsules. Adhering to this philosophy, we adopted the Wide residual-SE-based backend inspired by the recently introduced NF-Net or normalizer free network \cite{brock2021high}. Further, we incorporated an attention module into the backbone network, which yielded an accuracy of 95.35\% on the CIFAR-10 \cite{krizhevsky2009learning} dataset, which is an improvement of 5.6\% over the benchmark \cite{sabour2017dynamic} and 0.21\% improvement over the state-of-the-art \cite{tsai2020capsules}. However, the Wide residual-SE block was causing a significant increase in the trainable parameters. We adopted a Wide bottleneck residual block to curtail this, resulting in an improvement of 95.71\% from 95.35\%. 

Table \ref{tab4} depicts the architectural progression towards the proposed WideCaps architecture. In the $1^{st}$ two rows, we have listed the architectural details of Sabour \cite{sabour2017dynamic}, Zhao \cite{zhao2020efficient}, along with the performance achieved on CIFAR 10  \cite{krizhevsky2009learning} dataset. Unlike the methodology proposed by Zhao \cite{zhao2020efficient}, we adopted Wide ResNet instead of ResNet v2 in WideCaps v1 and achieved an accuracy of 93.20\% with a significant improvement of 3.49\%. In WideCaps v2, we adopted a modified FM routing for the purpose that is discussed in Section \ref{algo} and achieved an accuracy of 94.66\% with an improvement of 1.46\%. In WideCaps v3, we embedded Squeez and Excitation blocks and achieved a slight improvement of 0.20\%. However, WideCaps v3 resulted in computation cost; to curtail this, we adopted a wide bottleneck residual connection with SE blocks in WideCaps v4 and achieved an accuracy of 95.71\% (+0.36\% in comparison with WideCaps v3). In WideCaps v5, we adopted attention-based capsules to achieve state-of-the-art performance with an accuracy of 96.01\% on the CIFAR-10 dataset \cite{krizhevsky2009learning} and thus, we arrived at this configuration.
\section{Conclusion}
\label{sec_four}
This study presents a novel capsule network-based architecture called WideCaps for efficiently dealing with complex images for image classification. The proposed model couples the capabilities of CNN and the capsules to achieve the defined objective. WideCaps uses a wide bottleneck residual connection with squeeze and excitation attention block as the backbone network, followed by the attention capsules guided by the modified FM routing algorithm. The attention module expedited the flow of relevant features throughout the network by suppressing the counterpart. Notably, WideCaps achieved state-of-the-art results on CIFAR-10 and Fashion MNIST with highly competitive performance on the SVHN dataset. Our future work includes investigating the performance of the WideCaps on more complex datasets such as ImageNet and the quantitative investigation of the equivariance property.


\end{document}